\documentclass[letterpaper, 10 pt, conference]{ieeeconf}  %

\IEEEoverridecommandlockouts                              %

\let\labelindent\relax
\usepackage{enumitem} %

\usepackage{amsmath}
\usepackage{amssymb}
\usepackage{mathtools}
\usepackage{booktabs}
\usepackage[]{graphicx}
\usepackage[font=footnotesize]{caption}
\usepackage[font=footnotesize]{subcaption}
\usepackage{gensymb}
\usepackage{lipsum}
\usepackage{float}
\usepackage{color}
\usepackage{caption}
\usepackage{threeparttable}
\usepackage{todonotes}
\usepackage[]{algorithm2e}
\usepackage{siunitx}
\usepackage{hyperref}
\usepackage{cite}

\newcommand{\Sec}{Section~}

\newcommand{\Fig}{Fig.~}

\newcommand{\eg}{e.g., }
\newcommand{\ie}{i.e., }

\newcommand{\expm}{\exp}
\DeclareMathOperator{\Expm}{Exp}
\DeclareMathOperator{\Logm}{Log}

\global\long\def\se{\mathfrak{{se}}}
\global\long\def\SE{\text{{SE}}}

\global\long\def\T{\mathtt{{T}}}

\newcommand{\gT}{\bar{\T}}
\newcommand{\gA}{\bar{A}}
\newcommand{\N}{\mathcal{N}}

\newcommand{\cov}{\Sigma}
\newcommand{\gcov}{\bar\Sigma}
\newcommand{\bxi}{\pmb{\xi}}
\newcommand{\beps}{\pmb{\epsilon}}
\newcommand{\bsig}{\pmb{\sigma}}
\newcommand{\gbxi}{\pmb{\xi}}

\global\long\def\ki{k_{i}}

\newcommand{\RE}{\text{RE}}

\newcommand{\be}{\mathbf{e}}
\newcommand{\bmu}{\pmb{\mu}}
\newcommand{\Ad}{\text{Ad}}

\definecolor{somegray}{rgb}{0.5, 0.5, 0.5}
\newcommand{\darkgrayed}[1]{\textcolor{somegray}{#1}}
\makeatletter
\newcommand*\titleheader[1]{\gdef\@titleheader{#1}}
\AtBeginDocument{%
	\let\st@red@title\@title
	\def\@title{%
		\vskip-3em
		\bgroup\normalfont\large\centering\@titleheader\par\egroup
		\vskip1.5em\st@red@title}
}
\makeatother

\titleheader{\darkgrayed{This paper has been accepted for publication at the ICRA Workshop on Dataset Generation and Benchmarking of SLAM Algorithms for Robotics and VR/AR, Montreal, 2019.}}

\title{\LARGE \bf
	Rethinking Trajectory Evaluation for SLAM:\\
	a Probabilistic, Continuous-Time Approach
}

\author{Zichao Zhang, Davide Scaramuzza
	\thanks{The authors are with the Robotics and Perception Group, Dep. of Informatics, University of Zurich , and Dep. of Neuroinformatics, University of Zurich and ETH Zurich, Switzerland--- \url{http://rpg.ifi.uzh.ch.}}%
}

\begin{document}

\maketitle
\thispagestyle{empty}
\pagestyle{empty}

\begin{abstract}
Despite the existence of different error metrics for trajectory evaluation in SLAM, their theoretical justifications and connections are rarely studied, and few methods handle temporal association properly.
In this work, we propose to formulate the trajectory evaluation problem in a probabilistic, continuous-time framework.
By modeling the groundtruth as random variables, the concepts of absolute and relative error are generalized to be likelihood.
Moreover, the groundtruth is represented as a piecewise Gaussian Process in continuous-time.
Within this framework, we are able to establish theoretical connections between relative and absolute error metrics and handle temporal association in a principled manner.
\end{abstract}
\section{Introduction} \label{sec:intro}

Visual(-inertial) odometry (VO/VIO) and simultaneous localization and mapping (SLAM) are important building blocks in robotic systems, as they provide accurate state estimate for other tasks, such as control and planning.
To benchmark such algorithms, the most used method is to evaluate the estimated trajectory (\ie timestamped pose series) with respect to the groundtruth.

The central task for trajectory evaluation is to summarize certain metrics from the estimate and the groundtruth to indicate the performance.
There are many established evaluation methods, most notably the absolute trajectory error (ATE) \cite{Sturm12iros} and the relative error (RE) \cite{Geiger12cvpr}.
While these methods are widely used in practice and can be indicative of the performance, there are still many open problems.
Specifically, in this paper, we are interested in the following:
\begin{enumerate}
	\item It is well known that different metrics reflect different properties of estimate \cite{Zhang18iros}.
	However, the connection between them is not clear.
	Indeed, it is observed in practice that relative and absolute errors are often highly correlated (\eg \cite{Sturm12iros}),
	but no theoretical proof has been proposed before.
	\item Almost all the existing methods assume perfect temporal correspondences or adopt a naive matching strategy.
	For example, to find the corresponding groundtruth of the estimate at time $t$, most tools simply use the closest groundtruth, which is only acceptable when the groundtruth is of sufficiently high temporal resolution.
	There is no principled method currently to take into consideration of the imperfect temporal association, which can in practice have an impact for low rate groundtruth providers or missing data.
\end{enumerate}

In this work, we proposes to formulate the trajectory evaluation in a probabilistic, continuous-time framework.
First, instead of considering the groundtruth as deterministic values, we model it as random variables and thus generalize the concepts of absolute and relative error as likelihood.
While this step seems trivial (e.g., sum of squared error is simply the likelihood from Gaussian uncertainties), this allows us to draw connection between relative and absolute error (\Sec\ref{sec:traj_likelihood}).
Second, to reason about temporal association in a principled manner, we propose to use Gaussian Process (GP) to represent the groundtruth.
As a probabilistic and continuous-time representation, GP reports uncertainty for any query time, which, for example, gives higher uncertainties for query times far away from the actual groundtruth samples.
In this way, the effect of imperfect temporal association can be handled elegantly, which is not possible with many other continuous-time representations, such as polynomials.

It is well known that the trajectory evaluation problem is mainly complicated by the unobservable degrees-of-freedoms (DoFs) in the estimator \cite{Zhang18iros}, and thus is specific to different sensing modalities.
For simplicity, throughout the paper, we present the framework for trajectory estimates with unknown rigid-body transformations (\eg stereo or RGB-D sensors).
However, our method can be adapted to other interesting setups (monocular, visual-inertial) in future work.

\subsection{Contributions and Outline}
The contributions of this work are:
\begin{itemize}
	\item we provide the first theoretical proof that draws connection between relative and absolute error.
	\item the proposed probabilistic, continuous-time framework is the first that is able to handle temporal association properly.
\end{itemize}

The rest of the paper is structured as follows.
In \Sec\ref{sec:notations}, we formulate the evaluation problem of interest and introduce essential notations.
In \Sec\ref{sec:traj_likelihood}, the generalized versions of RE and ATE are derived with the probabilistic modeling of the groundtruth, and the connection between them are presented.
In \Sec\ref{sec:gp_gt}, we show how to represent the groundtruth as a Gaussian process on \SE(3).
Finally, we conclude the paper and discuss future work in \Sec\ref{sec:conclusion}. 
\section{The Trajectory Evaluation Problem} \label{sec:notations}
\subsection{Notations}
In this work, we are interested in evaluating the estimate consisting of 6 DoF poses, which is mostly common for VO/VIO/SLAM setups.
We parameterize a 6 DoF pose as an element (a $4\times4$ matrix) in matrix Lie group $\SE(3)$, denoted as $\T$.
As for the subscripts, $\T_{a, b}$ denotes the pose of frame $b$ with respect to frame $a$ (expressed in frame $a$).

To express the uncertainty associated with $\T$, we adopt the method in \cite{Barfoot14tro} to define distributions directly on the corresponding lie algebra $\se(3)$. Basically, to represent a stochastic pose $\T^{\prime}$, we use a deterministic pose ${\T}$ and a random variable $\bxi \in \mathbb{R}^6$. The stochastic pose is then
\begin{equation}
\T^{\prime} = {\T}\expm(\bxi^{\wedge}) 
\triangleq {\T}\Expm(\bxi),
\end{equation}
where $\exp(\cdot)$ is the exponential map of $\SE(3)$. For a Gaussian distribution $\bxi \sim \N(\mathbf{0}, \cov)$, we can write  $\T^{\prime} \sim \N_{\T^{\prime}}({\T}, \cov)$.

We will also heavily use the adjoint in $\SE(3)$, denoted as $\Ad_{\T}$ (a $6\times 6$ matrix), and some useful identities that will be used are:
\vspace{-1em}
\begin{eqnarray}
\T\Expm(\bxi) &=& \Expm(\Ad_{\T}\bxi)\T,
\label{eq:adj_change_order}\\
\det|\Ad_\T| &=& 1,
\label{eq:adj_det}\\
\Ad_{\T_1}\Ad_{\T_2} &=& \Ad_{\T_1\T_2},
\label{eq:adj_concate}
\end{eqnarray}
and \eqref{eq:adj_concate} also entails $(\Ad_\T)^{-1} = \Ad_{\T^{-1}}$ by setting $\T_1 = \T_2^{-1}$.

\subsection{Estimate, Groundtruth and Evaluation}
Technically speaking, a trajectory is a continuous function that maps from time to the state of interest (\eg 6 DoF pose, velocity).
In practice, we usually only have access to a limited number of samples from this continuous function.
Therefore, we can write the groundtruth and the estimate as
\begin{eqnarray}
\bar{X} = \{\gT_i\}_{i=1}^{N},\;\bar{\tau}=\{\bar{t}_i\}_{i=1}^{N}
\label{eq:gt}\\
X = \{\T_i\}_{i=1}^{M},\;\tau=\{t_i\}_{i=1}^{M}.
\label{eq:est}
\end{eqnarray}
Note that the times for the groundtruth and estimate $\bar{\tau}$  and $\tau$ are not necessarily the same.
For the case of perfect temporal association, we should have
$
\forall t_i \in \tau, \exists  \bar{t}_{\ki}\in\bar\tau:
\bar{t}_{\ki}=t_i, 
$
where $\ki$ is the index of the matching groundtruth for $t_i$.
In addition, we intentionally associate uncertainties with the groundtruth:
\begin{equation}
\bar\Lambda = \{\gcov_i\}_{i=1}^{N}\;\text{where}\; 
\gT^{\prime}_i \sim \N_{\gT^{\prime}_i}({\gT_i}, \gcov). \label{eq:gt_cov}
\end{equation}
One may argue that the uncertainties in the groundtruth are negligible.
While it is usually true for a single pose, we will see later that introducing uncertainties to groundtruth allows us to formulate the relative error in a more principled manner (\Sec\ref{subsec:gen_re}) and handle imperfect temporal association elegantly (\Sec\ref{sec:gp_gt}).

With \eqref{eq:gt}, \eqref{eq:est} and \eqref{eq:gt_cov} containing all the information about the estimated and groundtruth trajectories, a general performance metric should be a function of the form
\begin{equation}
f(\bar{X}, \bar\Lambda, \bar{\tau}, X, \tau).
\label{eq:perform_metric}
\end{equation}
The evaluation problem is about designing such a performance metric.
Next, we will generalize the concepts of ATE and RE with the aforementioned notations.
 
\begin{figure}
	\centering
	\includegraphics[width=0.9\linewidth]{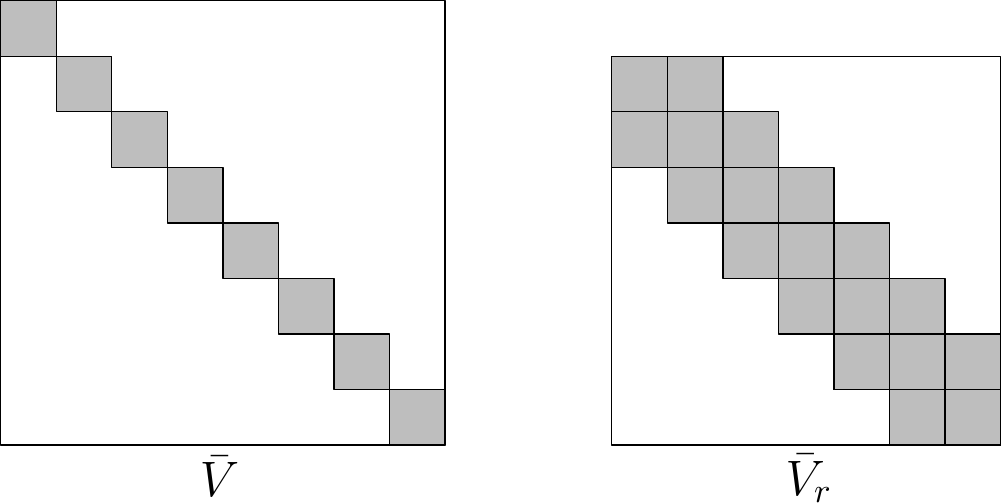}
	\caption{Block patterns of the covariance of groudtruth $\bar{X}$ and relative groudtruth $\bar{X}_r$.
	Note that the relative poses $\gT_{i, i+1}$ in $\bar{X}$ are actually correlated, as captured by the off-diagonal blocks in $\bar{V}_r$.
	}
	\label{fig:cov_gt_rel}
	\vspace{-0ex}
\end{figure}
\section{Generalized relative and absolute error} \label{sec:traj_likelihood}
In this section, we derive the counterparts of the commonly used RE and ATE from a probabilistic perspective, which can be seen as the generalized versions of them.
We assume perfect temporal association in this section, and the general case is handled in \Sec\ref{sec:gp_gt}.
To avoid the cluttering of subscripts, we denote the \textit{corresponding groundtruth and estimate} as
\begin{eqnarray}
\bar{X} = \{ \gT_{i} \}_{i=1}^{M},\;
\bar\Lambda = \{\bar{\cov}_i\}_{i=1}^{M}; \;
X = \{ \T_{i} \}_{i=1}^{M}.
\end{eqnarray}

\subsection{Generalized Relative Error}
\label{subsec:gen_re}
The basic idea of RE is to perform evaluation on the relative poses between pose pairs, and in this way, the ambiguity of the absolute reference frame is removed.
Here we consider the relative poses of consecutive estimated poses. To put it formally, we have the relative poses as
\begin{eqnarray}
\bar X_r &=& \{\gT_{i}^{-1}\gT_{i+1}\}_{i=1}^{M-1} =
\{\gT_{i, i+1}\}_{i=1}^{M-1}, \label{eq:rel_pose_gt}\\
X_r &=& \{\T_{i}^{-1}\T_{i+1}\}_{i=1}^{M-1} = 
\{\T_{i, i+1}\}_{i=1}^{M-1}.
\label{eq:rel_pose_est}
\end{eqnarray}
Since we treat the groundtruth as stochastic variables, its covariance also need to considered.
For the stochastic poses $\gT^{\prime}_{i}$ and $\gT^{\prime}_{i+1}$
\begin{eqnarray}
\gT^{\prime}_{i, i+1} &=& (\gT^{\prime}_{i})^{-1} (\gT^{\prime}_{i+1})
= (\gT_{i}\Expm(\gbxi_{i}))^{-1} \gT_{i+1}\Expm(\gbxi_{i+1}) \nonumber \\
&\overset{\eqref{eq:adj_change_order}}{=}& \gT_{i, i+1}\Expm(-\gA_{i, i+1}\gbxi_{i})\Expm(\gbxi_{i+1}),
\end{eqnarray}
where $\bar{A}_{i, j} \triangleq \Ad_{\gT_{i, j}}$ is the adjoint map of $\SE(3)$.
Then the random variable $\gbxi_{i, i+1}$ satisfies
\begin{equation}
\Expm(\gbxi_{i, i+1})  = \Expm(-\gA_{i, i+1}\gbxi_{i})\Expm(\gbxi_{i+1}),
\end{equation}
and $\gbxi_{i, i+1}$ can be expressed as infinite series using Baker-Campbell-Hausdorff (BCH) formula\cite{Barfoot14tro}.
Since both $\gbxi_i$ and $\gbxi_{i+1}$ are small (\ie the groundtruth is usually of high accuracy), we are justified to keep the first-order terms
\begin{equation}
\gbxi_{i, i+1} \simeq -\gA_{i, i+1}\gbxi_{i} + \gbxi_{i+1}.
\label{eq:rel_e_tfs}
\end{equation}
Stacking $\{\gbxi_i\}_{i=1}^{M}$ and $\{\gbxi_{i, i+1}\}_{i=1}^{M-1}$ as column vectors ${\bsig}$ and ${\bsig}_r$, we can write \eqref{eq:rel_e_tfs} collectively
$
\bsig_r = Q \bsig,
$
Since $\bsig \sim \N_{\bsig}(0, \bar{V})$ with $\bar{V} =\text{diag}(\gcov_i)$, we have \cite[p.~41]{Petersen12MatrixCookBook}
\begin{equation}
\bsig_r \sim \N_{\bsig_r}(0, Q\bar{V}Q^{\top}) = \N_{\bsig_r}(0, \bar{V}_r),
\label{eq:rel_error_pdf}
\end{equation}
where the block patterns in $\bar{V}$ and $\bar{V}_r$ are shown in \Fig\ref{fig:cov_gt_rel}.

With \eqref{eq:rel_pose_gt}, \eqref{eq:rel_pose_est} and \eqref{eq:rel_error_pdf}, we can now represent the error between $X_r$ and $\bar X_r$ as the probability $p(X_r | \bar X_r)$.
We first calculate the difference in the relative poses $\mathbf{e}_r$, which is a column vector stacked up by
\begin{eqnarray}
\be_i \triangleq \Logm((\gT_{i, i+1})^{-1} \T_{i, i+1});
\label{eq:rel_e_i}
\end{eqnarray}
then the probability of the estimate given the groundtruth, which is our \emph{generalized relative error}, is
\begin{eqnarray}
P_{\RE}(X | \bar{X}, \bar{\Lambda})
 = \N_{\mathbf{e}_r}(0, \bar{V}_r).
\label{eq:re_prob}
\end{eqnarray}
It is worth noting that, from \eqref{eq:rel_e_tfs}, we can see that there is actually correlation between $\bxi_{i-1, i}$ and $\bxi_{i, i+1}$, which is captured by the off-diagonal blocks in $\bar{V}_r$ (\Fig\ref{fig:cov_gt_rel}).

\begin{figure}
	\centering
	\includegraphics[width=0.9\linewidth]{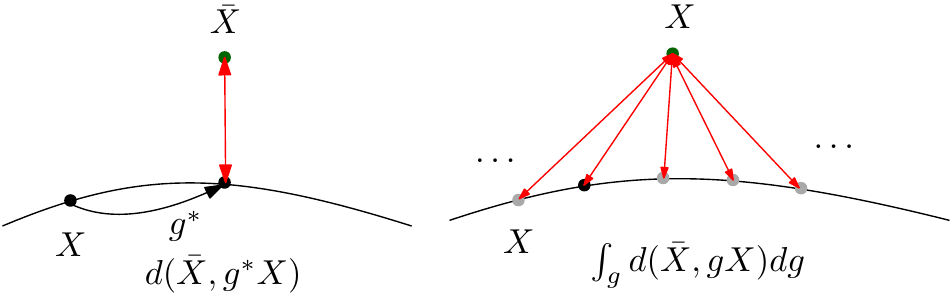}
	\caption{
	Illustration of conventional ATE and the generalized absolute error. The curve denote all the estimate that is equivalent to $X$ due to the unobservable DoFs \cite{Zhang18iros}.
	\textbf{Left}:
	in commonly used ATE, an optimal alignment transformation is first calculated, and the error is computed between the groundtruth and the transformed trajectory.
	\textbf{Right}:
	we propose to integrate over the equivalent parameters to eliminate the need of specifying a certain alignment transformation.
	}
	\label{fig:gen_ate}
	\vspace{-0ex}
\end{figure}

\subsection{Generalized Absolute Error}
\label{subsec:gen_ate}

Instead of eliminating the unknown transformation/reference frame $g$ by using the relative poses, ATE tries to find an optimal alignment transformation $g^{*}$ (usually in terms of position errors) first and then use it to calculate error metrics (in the world frame).
The approach is inherently problematic due to the fact that $g^{*}$ is calculated from the estimate and then used in turn to evaluate the estimate. In a probabilistic notation, the commonly used ATE is actually
\begin{equation}
p(X | \bar{X}, g)|_{g=g^{*}},
\label{eq:ate_problem}
\end{equation}
and worse still, the optimal $g^{*}$ is also not well-defined.

Can we get rid of $g$ in calculating the absolute error? Looking at \eqref{eq:ate_problem}, a straightforward idea is to marginalize $g$.
At first thought, marginalization seems to lose the information about the reference frame.
However, from the perspective of an estimator (whose performance we want to evaluate), there is simply \textit{no information} about the unobservable DoFs.
Therefore, we propose to marginalize $g$ in \eqref{eq:ate_problem} to get the \emph{generalized absolute error}, as illustrated in \Fig\ref{fig:gen_ate}.

In our setup, $g$ is a rigid-body transformation parameterized as an element in $\SE(3)$. Similar to \cite{Barfoot14tro}, we integrate exponential coordinates $\bxi_{g} \in \mathbb{R}^6$ directly :
\begin{equation}
L_{\text{AE}} (X|\bar{X}, \bar{\Lambda}) = \int_{\mathbb{R}^6} \prod_{i=1}^{M}
\mathcal{N}_{\beps_{i}}(0, \gcov_i) d\bxi_g,
\label{eq:gen_ate_inte}
\end{equation}
where $\beps_{i}$ is the difference for each pose
\begin{equation}
\beps_{i} = \Logm(\gT_{i}^{-1}  g  \T_{i}) = \Logm(\gT_{i}^{-1}  \Expm(\bxi_{g})  \T_{i}).
\label{eq:gen_ate_epsi}
\end{equation}
Note that we are using an noninformative prior \cite[p.~117]{Bishop06book}, namely $p(\bxi_g) = 1,\; \forall \bxi_g \in \mathbb{R}^6$,
in \eqref{eq:gen_ate_inte}, which is not a proper distribution.
This means that $L_{\text{AE}}$ is \emph{not} a valid PDF of the estimate $X$ any more.

Unfortunately, \eqref{eq:gen_ate_inte} is non-trivial to calculate analytically.
$\beps_{i}$ in \eqref{eq:gen_ate_epsi} can be written as infinite series (BCH formula).
But since we are considering $\bxi_{g} \in \mathbb{R}^6$ (\ie not necessarily small), we cannot keep the first-order terms only.
However, by examining the first-order terms, we are able to draw connections between \eqref{eq:re_prob} and \eqref{eq:gen_ate_inte}.

\subsection{Connection between Absolute and Relative Error}
\label{subsec:ae_eq_re}
Conventional ATE and RE reason about estimation error in completely different ways.
However, it is usually observed that they are strongly correlated \cite{Sturm12iros}.
With our probabilistic setup, it is actually possible to establish the connection between generalized absolute and relative error theoretically.

Starting from the generalized absolute error \eqref{eq:gen_ate_inte},
without loss of generality, we assume that $\gT_1 = \T_1$.
The reason is that we can always find a $\Delta\T$ such that $\gT_1 = \hat{\T}_1 \triangleq\Delta\T\cdot\T_1$,
and using $\hat{\T}_i$ for evaluation should give the same error metric as  $\T_i$, since they are equivalent estimate\cite{Zhang18iros}.
Under this assumption, \eqref{eq:gen_ate_epsi} can be simplified for the first pose as 
\begin{equation}
\beps_1 = 
\Logm(  \Expm(\gA_1^{-1} \bxi_{g})\gT_{1}^{-1}  \T_{i})  =
\gA_{1}^{-1}\bxi_g,
\label{eq:gen_ate_init_pose} 
\\
\end{equation}
which also gives
\begin{equation}
d\beps_{1} = \det|\gA_{1}^{-1}|d\bxi_{g} \overset{\eqref{eq:adj_det}}{=} d\bxi_{g}.
\label{eq:deps_dg}
\end{equation}
We then further write the rest of the PDFs (\ie $i>1$) in \eqref{eq:gen_ate_inte} in terms of $\beps_{1}$ by repeatedly applying \eqref{eq:adj_change_order} and \eqref{eq:adj_concate} and keeping the first-order terms in BCH formula.
For instance, for the second pose:
\begin{eqnarray}
\beps_2 &=& \Logm(\gT_{12}^{-1} \underbrace{\gT_{1}^{-1}\Expm(g)\T_{1}}_{\beps_1}
\T_{12}) \nonumber\\
&=& \Logm(\Expm(\gA_{21}\beps_i)\gT_{12}^{-1} \T_{12})
\simeq \gA_{21}\beps_1 + \mathbf{e}_1. \nonumber
\end{eqnarray}
And in general, for $h \in [1, M]$
\begin{equation}
\beps_{h} \simeq \gA_{h1}\beps_{1} + 
\gA_{h2}\be_1 + \cdots 
+ \gA_{h, h-1}\be_{h-2} + \be_{h-1},
\label{eq:eps_approx}
\end{equation}
where $\be_i$ is the relative difference defined in \eqref{eq:rel_e_i}.
With the approximation \eqref{eq:eps_approx}, the individual Gaussian PDF can be written in terms of $\beps_1$ as \cite[p.~41]{Petersen12MatrixCookBook}
\begin{equation}
\mathcal{N}_{\beps_{h}}(0, \gcov_h) \simeq
\frac{1}{\det|\gA_{h1}|} \mathcal{N}_{\beps_{1}}(\bmu_h, V_h)= \mathcal{N}_{\beps_{1}}(\bmu_h, V_h),
\label{eq:approx_ate_pdf}
\end{equation}
where $\bmu_1 =\mathbf{0}, V_1 = \gcov_1$ and for $h = 2, 3, \cdots, M$:
\begin{eqnarray}
\bmu_h &=& -\gA_{12}\be_1 -\gA_{13}\be_2 - \cdots \gA_{1h}\be_{h-1}
\label{eq:mu_abs}\\
V_h &=& \gA_{1h}\gcov_{h}\gA_{1h}^{\top}.
\label{eq:var_abs}
\end{eqnarray}
Finally, plugging \eqref{eq:deps_dg} and  \eqref{eq:approx_ate_pdf} into \eqref{eq:gen_ate_inte}, we have the following approximation
\begin{equation}
L_{\text{AE}} \simeq \tilde{L}_{\text{AE}} = 
\int_{\mathbb{R}^6} 
\prod_{i=1}^{M}\mathcal{N}_{\beps_{1}}(\bmu_h, V_h) d\beps_1.
\label{eq:gen_ate_approx}
\end{equation}
Using the fact that the product of multivariate Gaussian PDFs is a scaled multivariate Gaussian PDF \cite{Bromiley03tina}, \eqref{eq:gen_ate_approx} becomes
\begin{equation}
\tilde{L}_{\text{AE}} = 
S_{\text{AE}} \int_{\mathbb{R}^6} 
\mathcal{N}_{\beps_1}(\bmu_s, V_s) d\beps_1 
= S_{\text{AE}},
\end{equation}
where $S_{\text{AE}}$ is again a Gaussian function.
It is worth noting that $S_{\text{AE}}$ only depends on relative poses via \eqref{eq:mu_abs} and \eqref{eq:var_abs} but not the alignment transformation $\bxi_g$.
It is possible to further prove  that
\begin{equation}
S_{\text{AE}}(X | \bar{X}, \bar{\Lambda}) = P_{\RE}(X | \bar{X}, \bar{\Lambda}).
\label{eq:equiv_ate_re}
\end{equation}
The proof is straightforward (\eg comparing the covariance matrix and quadratic terms in the Gaussian functions) but lengthy, and thus omitted for the sake of space.
In words, under our probabilistic setup, the generalized absolute error \eqref{eq:gen_ate_inte} is the same as the generalized relative error \eqref{eq:re_prob} up to the first order.
\subsection{Discussion}
To summarize, in this section, we generalize the concepts of RE and ATE respectively to \eqref{eq:gen_ate_inte} and \eqref{eq:re_prob} from a probabilistic perspective. An interesting observation is that up to the first-order terms, the generalized relative error and absolute error is actually the same.
Admittedly, the equivalence that we established in \Sec\ref{subsec:ae_eq_re} is only valid for first-order terms.
One future direction is to examine higher order terms in details.
 
\begin{figure}
	\centering
	\includegraphics[width=0.9\linewidth]{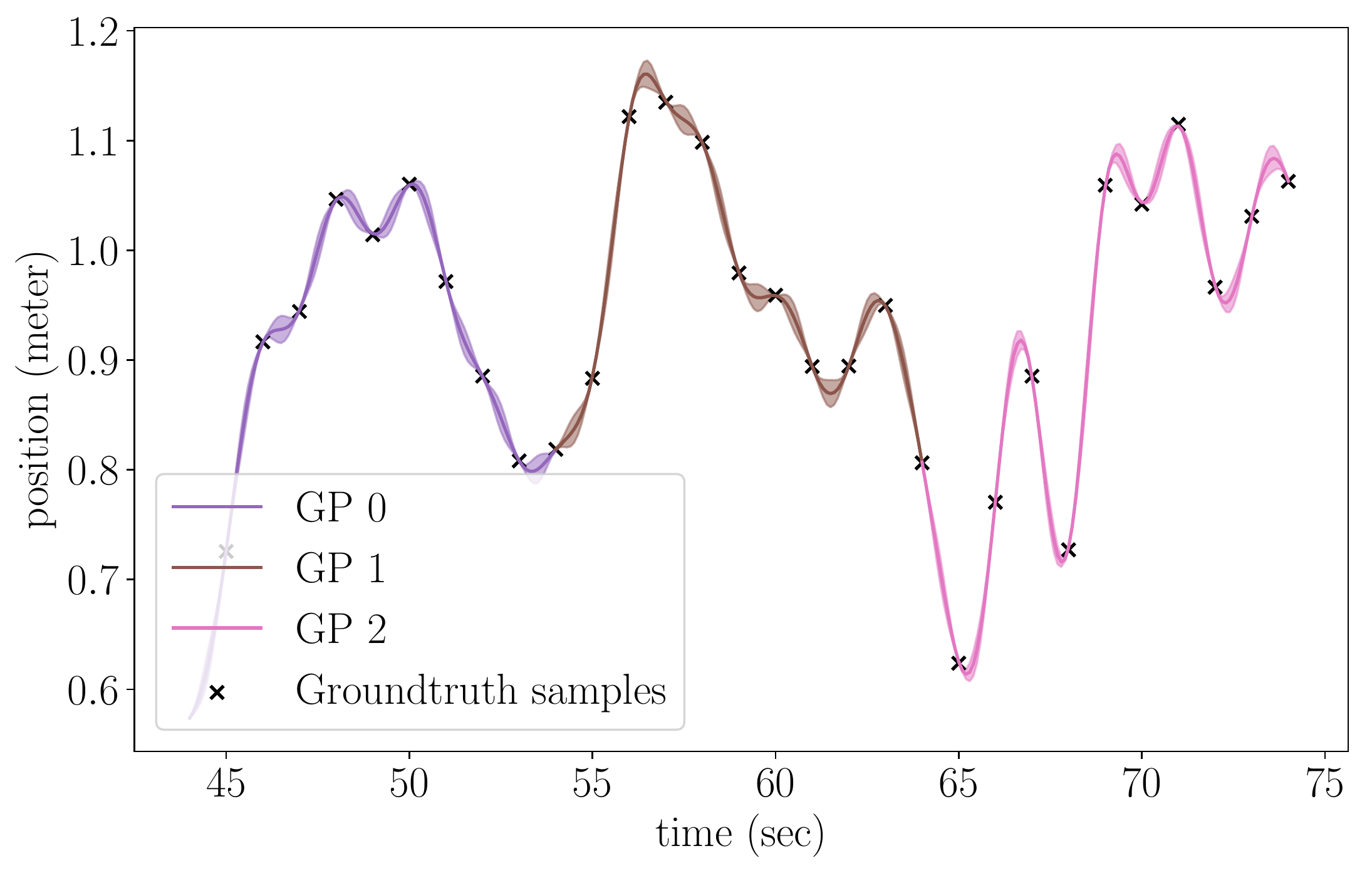}
	\caption{
	Example of using a piecewise Gaussian process to represent groundtruth.
	In this example, three GPs are used to represent a position component.
	The shaded area represents the uncertainty reported by the Gaussian process regression.
	It can be seen that the regressed result fits the actual samples well, and uncertainties increase with the distance from the groundtruth samples.
	}
	\label{fig:gp_example}
	\vspace{-0ex}
\end{figure}

\section{Groundtruth as a Gaussian Process} \label{sec:gp_gt}
In this section, we will first give a brief introduction of GP regression and then describe how to use GP to model groundtruth trajectories in a continuous-time and probabilistic manner.

\subsection{Gaussian Process Regression}
A Gaussian process is a collection of random variables, and any subset of them has a joint Gaussian distribution \cite{Rasmussen05gpml}.
In the context of a regression task, suppose we know the samples at 
$\mathbf{z} = \{z_i\}_{i=1}^{P}$ with the output $\mathbf{y} = \{y_i\}_{i=1}^{P}$,
and we would like to know the output value $y^*$ at $z^*$.
Under the assumption of Gaussian process, we have
\begin{equation}
\begin{bmatrix}
\mathbf{y} \\
y^*
\end{bmatrix}
\sim
\mathcal{N}(\mathbf{0},
\begin{bmatrix}
K_{\mathbf{z}\mathbf{z}}& K_{\mathbf{z}z^*}\\
K_{{z^*}\mathbf{z}}& k(z^*, z^*)
\end{bmatrix}
),
\label{eq:gp_prior}
\end{equation}
where $K_{\mathbf{z}\mathbf{z}}^{i, j} = k(z_i, z_j)$, 
$K_{z^*\mathbf{z}}^{i} = k(z^*, z_i)$ and $K_{\mathbf{z}z^*}^{i} = k(z_i, z^*)$.
Then the GP regression simply takes the conditional distribution
\begin{equation}
y^* \sim \mathcal{N}(
K_{{z^*}\mathbf{z}} K_{\mathbf{z}\mathbf{z}}^{-1} \mathbf{y},
k(z^*, z^*) -
K_{{z^*}\mathbf{z}} K_{\mathbf{z}\mathbf{z}}^{-1} K_{\mathbf{z}{z^*}}),
\label{eq:gp_regress}
\end{equation}
which gives both the regressed value and variance.

Obviously, the properties of the prior \eqref{eq:gp_prior} and the regressed result \eqref{eq:gp_regress} depends on the function $k(\cdot)$.
$k(a, b)$ is called the \textit{kernel function}, and intuitively encodes the correlation of the outputs at $a, b$.
Often $k(\cdot)$ is a parameterized functions, whose parameters are the \textit{hyperparameters} of a GP.
Perhaps the most used kernel function is the Squared Exponential function
\begin{equation}
k_{\text{SE}}(a, b) =  \sigma^2 \exp(-\frac{(a-b)^2}{2l^2}),
\end{equation}
where $\sigma$ and $l$ are the hyperparameters.

GP is a flexible model that finds many applications in robotics (\eg motion planning \cite{dong2016motion}, state estimation \cite{Barfoot14rss}).
For simplicity, the above introduction is limited to the case where both the output and input are scalars. However, GP can also be generalized to vector input and output.
For a thorough description of GP (\eg optimization of hyperparameters), we refer the reader to \cite{Rasmussen05gpml}.

\subsection{Piecewise GP on $\SE(3)$}
The application of GP to 6 DoF poses is more complicated due to the rotation components in rigid-body poses.
We use a piecewise GP, similar to \cite{dong2017sparse}.
In particular, we divide the whole trajectory into several segments.
Within each segments, we select a reference pose $\T_{\text{ref}}$ and denote the poses inside this segment using the elements in $\se(3)$ around $\T_{\text{ref}}$, which can be seen as a vector space locally and expressed using a GP.

Using a piecewise GP, however, brings complication to the choice of hyperparameters.
Specifically, if we optimize a GP for each segment of the trajectory separately, there is no guarantee that the uncertainties and the mean value is continuous at the boundaries.
In practice, we adopt the following strategy: 
i) we select the segments so that the adjacent segments overlap (\eg 50\%) with each other;
ii) we use the same hyperparameters for all the segments.
An example of the resulting piecewise GP is illustrated in \Fig\ref{fig:gp_example}.

\subsection{Using GP in Trajectory Evaluation}
Once we have constructed the piecewise GP to represent the groundtruth, we are able to query the groundtruth at any given time, with uncertainty estimate using \eqref{eq:gp_regress}, which can then be directly used in our evaluation setup \eqref{eq:perform_metric}.

An example of the advantage of using GP can be observed in \Fig\ref{fig:gp_example}. 
We can see that for query time that is far from the groundtruth samples, the uncertainty increases.
Intuitively, with a larger uncertainty, the same difference in the mean value will result in a lower likelihood (\ie larger error) in \eqref{eq:gen_ate_approx} or \eqref{eq:re_prob}.
In this way, the temporal association is taken into account properly.

\section{Conclusion and Future Work}\label{sec:conclusion}
In this paper, we propose to formulate the trajectory evaluation problem in a probabilistic, continuous-time framework.
By using Gaussian process as the underlying representation and formulating the estimation error probabilistically, we are able to draw theoretical connections between relative and absolute error metrics and handle temporal association in a principled manner.

Future work could go in several aspects:
\begin{itemize}
	\item The equivalence of relative and absolute error metrics in \Sec\ref{sec:traj_likelihood} is only proved up to first order.
	It is interesting to know whether similar conclusions can be drawn for higher order terms.
	\item In \Sec\ref{sec:gp_gt}, to make the piecewise GP continuous (the mean values and variances), we adopted the simple strategy of using overlapping segments with shared hyperparameters. Ideally, the continuity should be enforced via additional constraints in the optimization of the hyperparameters.
	\item Thorough comparative experiments are needed to better understand the properties of the proposed method (compared to existing approaches) in different situations.
\end{itemize}

\bibliographystyle{IEEEtran}
\bibliography{../rpg_bib/all,references}

\end{document}